\documentclass[12pt]{article}
\usepackage{marvosym}
\usepackage{amsmath} % AMS Math Package
\usepackage{amsthm} % Theorem Formatting
\usepackage{amssymb}    % Math symbols such as \mathbb
\usepackage{graphicx} % Allows for eps images
\usepackage[dvips,letterpaper,margin=1in,bottom=0.7in]{geometry}
\usepackage{tensor}
\usepackage{calc}
\usepackage{tikz}

\usepackage{amsmath}
\makeatletter % Need for anything that contains an @ command 
\renewcommand{\maketitle} % Redefine maketitle to conserve space
{ \begingroup \vskip 10pt \begin{center} \huge {\bf \@title}
    \vskip 10pt \large \@author \hskip 20pt \@date \end{center}
  \vskip 10pt \endgroup \setcounter{footnote}{0} }
\makeatother % End of region containing @ commands
\usepackage{enumerate}
\let\baraccent=\= % rename builtin command \= to \baraccent
\renewcommand{\=}[1]{\stackrel{#1}{=}} % for putting numbers above =

\providecommand{\bd}[1]{\textbf{#1}}
\providecommand{\sct}[1]{\section{#1}}
\providecommand{\sbs}[1]{\subsection{#1}}

\providecommand{\be}[1]{\begin{enumerate}[#1]}

\providecommand{\ee}{\end{enumerate}}
\providecommand{\bq}{\begin{quote}}
\providecommand{\eq}{\end{quote}}

\usepackage{cancel}

\theoremstyle{definition}

\newbox\gnBoxA
\newdimen\gnCornerHgt
\setbox\gnBoxA=\hbox{$\ulcorner$}
\global\gnCornerHgt=\ht\gnBoxA
\newdimen\gnArgHgt
\def\gd #1{%
\setbox\gnBoxA=\hbox{$#1$}%
\gnArgHgt=\ht\gnBoxA%
\ifnum     \gnArgHgt<\gnCornerHgt \gnArgHgt=0pt%
\else \advance \gnArgHgt by -\gnCornerHgt%
\fi \raise\gnArgHgt\hbox{$\ulcorner$} \box\gnBoxA %
\raise\gnArgHgt\hbox{$\urcorner$}}
\providecommand{\beg}{\begin{enumerate}[I]}
\providecommand{\bee}{\begin{enumerate}[i]}

\providecommand{\ee}{\end{enumerate}}

\usepackage{forest}

\title{Instrumental convergence and power-seeking}
\author{David Thorstad}
\date{}

\usepackage{txfonts}
\usepackage{pxfonts}
\usepackage{url}
\usepackage{booktabs}
\usepackage{wrapfig}

\newcounter{numbcounter}
\makeatletter
\renewcommand\@biblabel[1]{}

\makeatother

\usepackage{setspace}

\usepackage{natbib}
\setcitestyle{aysep={}} 

\usepackage{float}
\usetikzlibrary{trees}
\usepackage{xurl}

\usepackage{setspace}

\begin{document}

\sloppy

\maketitle

\begin{abstract} \noindent  Recent years have seen increasing concern that artificial intelligence may soon pose an existential risk to humanity. One leading ground for concern is that artificial agents may be power-seeking, aiming to acquire power and in the process disempowering humanity. I show how the argument from power-seeking rests on a strong version of a claim known as the instrumental convergence thesis. I explore leading defenses of the instrumental convergence thesis and argue that none establishes the thesis in a strong enough form to ground the argument from power-seeking. I discuss implications for longtermism, the governance of artificial intelligence, and the methodology of studying risks posed by artificial agents.
 \end{abstract}

\doublespacing

\sct{Introduction} \label{Introduction}

Recent years have seen increasing concern that artificial intelligence may soon pose an existential risk to humanity. Significant concerns have been expressed by artificial intelligence pioneers such as Yoshua Bengio \citeyearpar{Bengio2023}, Geoffrey Hinton \citeyearpar{Metz2023}, and Stuart Russell \citeyearpar{Russell2019}. Leading artificial intelligence researchers have signed statements \citep{CAIS-Statement,Pause-Letter} calling for increased attention to existential risks, and many express sympathy for risk claims in expert surveys \citep{Grace2016,Grace2022,Muller2016,Zhang2022}. A raft of organizations have devoted significant resources to studying and mitigating existential risks from artificial intelligence.\footnote{These include nonprofits such as the Center for AI Safety and the Center for the Governance of AI; government institutes such as the UK AI Safety Institute and the US AI Safety Institute; frontier AI laboratories such as OpenAI and Anthropic; grantmakers such as Open Philanthropy and the Future of Life Institute; dedicated laboratories such as Conjecture and Redwood Research; and academic centers such as the Stanford Center for AI Safety and the CMU Safe AI Lab.} Concerns about existential risk are defended at book length by leading scholars \citep{Bostrom2014,Russell2019}, in policy reports \citep{Carlsmith2021,Cotra2020}, and in academic papers \citep{Bales2024,Bostrom2012,TurnerOptimalPolicies}.

One prominent ground for concern is that artificial agents may be power-seeking, aiming to acquire power and in the process disempowering humanity in a permanent and catastrophic fashion \citep{Bostrom2012,Carlsmith2021,Dung2025,Ngo2023}. Typically, concerns about power-seeking are rooted in the idea that power is an instrumentally convergent goal, roughly in the sense that a wide variety of agents will find power conducive to achieving their goals and hence will pursue power in order to achieve their goals \citep{Bostrom2012,Omohundro2008}.

My aim in this paper is to clarify the instrumental convergence thesis used to drive concerns about existential risk from power-seeking artificial intelligence and to argue that leading defenses of instrumental convergence do not establish the needed claim. Section \ref{PowerSeekingAndIC} presents the \textit{argument from power-seeking} used to argue that power-seeking artificial intelligence poses a significant existential risk, then clarifies the role of instrumental convergence in the argument from power-seeking. Section \ref{ArgFromMisalignment} explores a leading informal defense of instrumental convergence, the \textit{argument from misalignment}, and argues that it does not establish an instrumental convergence claim strong enough to drive the argument from power-seeking. Section \ref{Orbital-Model} explores a range of power-seeking theorems which aim to formally establish instrumental convergence results, focusing on the Orbital Markov Model of Alexander Turner and colleagues \citeyearpar{TurnerOptimalPolicies}. Section \ref{Orbital-Markov-Challenges} argues that power-seeking theorems likewise fail to establish the needed version of instrumental convergence. Section \ref{Discussion} concludes by discussing implications for the methodology of studying risks posed by artificial agents (Sections \ref{ClarifyPower}-\ref{NothingNothing}), the governance of artificial intelligence (Section \ref{PolicyImplications}) and longtermism (Section \ref{Longtermism}).

\sct{Power-seeking and instrumental convergence} \label{PowerSeekingAndIC}

The argument from power-seeking claims that artificial agents with a wide variety of goals will be motivated to seek power, thereby disempowering humanity and causing an existential catastrophe \citep{Bostrom2014,Carlsmith2021,Carlsmith-ChapterVersion,Ngo2023,TurnerOptimalPolicies}. Many formulations of the argument are possible, but here is a leading formulation due to Joe Carlsmith \citep{Carlsmith2021,Carlsmith-ChapterVersion}.\footnote{See \citet{Dung2025} for a related presentation of the argument from power-seeking.} Carlsmith holds that by 2070:\footnote{This argument is taken directly from Carlsmith \citeyearpar{Carlsmith-ChapterVersion}, with two modifications. First, I treat the premises as unconditional claims, whereas Carlsmith conditionalizes each premise on the previous premises. Second, I have added descriptive labels to each premise. A slightly expanded version of this argument can be found in \citep{Carlsmith2021}.}

\begin{quote}
\bd{(Possibility)} It will become possible and financially feasible to build relevantly powerful and agentic AI systems.
\end{quote}

\begin{quote}
\bd{(Incentives)} There will be strong incentives to do so.
\end{quote}

\begin{quote}
\bd{(Alignment Difficulty)} It will be much harder to build aligned (and relatively powerful and agentic) AI systems than to build misaligned (and relevantly powerful and agentic) AI systems that are still superficially attractive to deploy.
\end{quote}

\begin{quote}
\bd{(Power-Seeking)} Some such misaligned systems will seek power over humans in high-impact ways.
\end{quote}

\begin{quote}
\bd{(Disempowerment)} This problem will scale to the full disempowerment of humanity.
\end{quote}

\begin{quote}
\bd{(Catastrophe)} Such disempowerment will constitute an existential catastrophe.
\end{quote}

\noindent There are many ways to push back against the argument from power-seeking. We might raise technological challenges to Possibility, questioning the technological feasibility of constructing systems powerful enough to disempower humanity by 2070 \citep{Landgrebe2022,ThorstadAgainstSing}. We might unpack the different notions of disempowerment involved in Disempowerment and question whether the most problematic will come to pass \citep{Bales-Disempowerment}. Or we might deny Catastrophe, holding that a future without humanity would not be catastrophic, for example because the world is not made better by improving the lives of individuals who would otherwise not exist \citep{Narveson1973,Frick2017}, because our descendants might suffer \citep{Benatar2006}, or because our posthuman replacements might be wiser and more numerous than us \citep{Armstrong2013,Greaves2021}.

This paper pursues a different route. Leading arguments for Alignment Difficulty and Power-Seeking appeal to the idea that power is an instrumentally convergent goal \citep{Bostrom2014,Carlsmith2021}. In rough outline, Power-Seeking is defended on the grounds that power is valuable to agents with many different goals, and Alignment Difficulty is defended on the grounds that it is difficult to identify useful goals for which power would not be valuable. I want to challenge this appeal to instrumental convergence.

What exactly does instrumental convergence hold? A leading statement of instrumental convergence is due to Nick Bostrom:

\begin{quote}
\bd{(IC-B)} Several instrumental values can be identified which are convergent in the sense that their attainment would increase the chances of the agent's goal being realized for a wide range of final goals and a wide range of situations, implying that these instrumental values are likely to be pursued by many intelligent agents. \citep[p. 76]{Bostrom2012}
\end{quote}

\noindent IC-B contains an inference between two claims that we may have reason to treat separately \citep{GallowDivergence,ThorstadCarlsmithBlog}:

\begin{quote}
\bd{(Goal Realization)} There are several values which would increase the chances of an agent's final goal being realized, for a wide range of goals and a wide range of situations.
\end{quote}

\begin{quote}
\bd{(Goal Pursuit)} There are several values which would be likely to be pursued by a wide range of intelligent agents.
\end{quote}

\noindent IC-B asserts both Goal Realization and that Goal Realization implies Goal Pursuit. Many subsequent statements of instrumental convergence likewise treat instrumental convergence as Goal Realization, assuming or arguing that Goal Realization implies Goal Pursuit.\footnote{For example, Adam Bales and colleagues take instrumental convergence to be ``the claim that certain resource-acquiring, self-improving and shutdown-resisting subgoals are useful for achieving a wide variety of final goals'' \citep[p. 5]{Bales2024}. Leonard Dung states instrumental convergence as the claim that ``there are certain goals which are instrumentally useful for a wide range of final goals and a wide range of situations'' \citep[p. 10]{Dung2025}. And Richard Ngo and Adam Bales \citeyearpar{Ngo2023} adopt a formulation they attribute to Bostrom on which instrumental convergence ``states that some instrumental goals -- like survival, resource acquisition, and technological development -- are instrumentally useful for achieving almost any final goal.''}

Establishing Goal Pursuit may be more difficult than establishing Goal Realization for several reasons. One challenge that will not be pursued here is that many existing arguments from Goal Realization to Goal Pursuit assume that artificial agents are well-modeled as having and optimizing goals, often in something like the sense of expected-utility maximization. That may not be obvious \citep{Bales2025}.

The challenge that I want to pursue is that Goal Pursuit differs from Goal Realization in speaking of agents rather than goals, and of what agents will do rather than what would increase the chance of their goals being realized. This makes Goal Pursuit much harder to demonstrate, since agents have multiple goals and are not always willing to pursue one goal at the expense of all others. For example, there is no doubt that money is conducive to the achievement of many goals that I have. However, it does not follow that I would rob a bank tomorrow if I could get away with it. That is not because I have no use for the money, but rather because I also value the welfare of others, fairness and the rule of law. While I may be willing to bend these scruples from time to time, I am not willing to toss them dramatically aside, even for great instrumental gain. In the same way, what must be shown is not just that artificial agents would find power greatly conducive to many of their goals, but also that they will be so utterly unconcerned with the consequences that they find the complete and existentially catastrophic disempowerment of humanity to be an acceptable sacrifice in exchange for power. 

This last claim reminds us that even Goal Pursuit is not enough to ground the argument from power-seeking, since it says nothing about the degree of power that is likely to be pursued. To ground Disempowerment and Catastrophe, the argument from power-seeking needs to claim:

\begin{quote}
\bd{(Catastrophic Goal Pursuit)} There are several values which would be likely to be pursued by a wide range of intelligent agents to a degree that, if successful, would lead to the permanent and existentially catastrophic disempowerment of humanity.
\end{quote}

\noindent Catastrophic Goal Pursuit is a much stronger claim than Goal Pursuit. Most of us sometimes pursue money and other forms of power. Indeed, I very much hope to be paid monthly for my work. Many fewer of us pursue great power at significant expense to others, for example by robbing a bank. And precious few pursue global power, seeking total and permanent control over humanity. That is not just because we think we would not be successful but also, for most normal humans, because we count the prospect of world domination as rather unappealing.

Catastrophic Goal Pursuit is a strong claim, and it should be given a correspondingly strong argument. In the next three sections, I explore informal (Section \ref{ArgFromMisalignment}) and formal (Sections \ref{Orbital-Model}-\ref{Orbital-Markov-Challenges}) arguments for Catastrophic Goal Pursuit and argue that they do not succeed. In each case, I focus on a leading formulation of the argument that will be representative of many, though perhaps not all arguments for Catastrophic Goal Pursuit. The conclusion that follows will then be, at the very least, that these leading arguments do not establish Catastrophic Goal Pursuit.

\sct{The argument from misalignment} \label{ArgFromMisalignment}

One of the most common informal arguments for Catastrophic Goal Pursuit is what we might call the argument from misalignment \citep{Dung2025,Carlsmith2021,Ngo2023}. This argument is embedded in Carlsmith's formulation of the argument from power-seeking, which argues first for Alignment Difficulty and then on this basis for Power-Seeking. The argument goes roughly as follows:

\beg 
\item[] \bd{(Misaligned Deployment)} Humanity will soon develop and deploy a wide range of misaligned artificial agents.
\item[] \bd{(Misaligned Disempowerment)} There are several values which misaligned artificial agents would likely pursue to a degree that, if successful, would lead to the permanent and existentially catastrophic disempowerment of humanity.
\item[$\therefore$] \bd{(Catastrophic Goal Pursuit)} There are several values which would be likely to be pursued by a wide range of intelligent agents to a degree that, if successful, would lead to the permanent and existentially catastrophic disempowerment of humanity.
\ee

\noindent We can see why skeptics of Catastrophic Goal Pursuit remain unconvinced by evaluating both premises of the argument from misalignment.

To unpack the contents of Misaligned Deployment, we need to say what it means for artificial agents to be misaligned. Some authors understand misalignment as a property of agents' goals -- for example, Richard Ngo and Adam Bales \citeyearpar{Ngo2023} hold that ``goals [are] misaligned if they are undesirable from a human perspective.'' Others understand misalignment as a property of behavior -- for example, Joe Carlsmith \citeyearpar{Carlsmith-ChapterVersion} defines misaligned behavior as ``unintended behavior that arises specifically in virtue of problems with an AI system's objectives.'' Others are more pluralistic: for example, Leonard Dung defines an aligned model as one that ``pursues goals, has values or optimizes an objective function which correspond to the goals, values or objective function its desires want it to have'' and defines a misaligned model to be one that is not aligned \citep[p. 8]{Dung2025}. What is common to these definitions is the idea that misalignment involves goals or behaviors that are unintended or undesirable from a human perspective. For simplicity, I discuss misalignment as a property of the desirability of goals, though much of the discussion should carry over to other definitions. 

On this definition, it would not be surprising for artificial agents to be misaligned to some degree. A system which placed only slightly too much value on situations in which humans drink green tea would be misaligned in virtue of holding an undesirable goal. If Misaligned Deployment is meant to support Misaligned Disempowerment, we will need to understand misalignment in a much stronger sense. For example, we might understand misalignment as \textit{catastrophic underweighting}, in which agents assign significantly less disvalue to some types of human disempowerment than it would be desirable for them to assign. Alternatively, we might understand misalignment as \textit{rapid swamping}, in which agents value some goals to the extent that they could value feasible levels of goal achievement strongly enough to outweigh the disvalue of human disempowerment. What could be said in favor of thinking that such dramatically misaligned agents will be developed and deployed?

One concern is \textit{reward mis-specification} \citep{Pen2022,Ribero2020,Russell2019}. Since we cannot precisely formalize our values, we often reward agents during training for proxy goals that can come apart from what we care about. Stewart Russell offers the following illustration: 

\begin{quote}
Suppose we ask some future superintelligent system to pursue the noble goal of finding a cure for cancer -- ideally as quickly as possible, because someone dies from cancer every 3.5 seconds. Within hours, the AI system has read the entire biomedical literature and hypothesized millions of potentially effective but previously untested chemical compounds. Within weeks, it has induced multiple tumors of different kinds in every living human being so as to carry out medical trials of these compounds, this being the fastest way to find a cure. \citep[p. 138]{Russell2019}
\end{quote}

\noindent More immediate examples of reward mis-specification are ready to hand. For example, an agent rewarded for minimizing the number of visibly un-watered tomatoes learned to place a bucket over its head and stop watering tomatoes \citep{Leike2017}. Similarly, a system trained to stack a red lego on top of a blue lego was rewarded for the height of the top of the red lego when released \citep{Krakovna2020}. The system learned to rotate the red lego to stand vertically, then let it fall.

Another concern is \textit{goal misgeneralization}: goals that lead agents to perform well during training may not lead to good performance in novel situations \citep{Ngo2023,Langosco2022,Shah2022}. Examples of goal misgeneralization abound. For example, an agent trained to open chests with keys developed the goal of collecting keys alongside the goal of opening chests \citep{Langosco2022}. In key-rich environments, the agent then took to collecting keys and ignoring chests. Likewise, an agent trained to harvest trees learned from tree-rich environments to chop trees as quickly as possible, leading to deforestation in novel environments \citep{Shah2022}.

A final concern is the difficulty of detecting misaligned goals. Detecting misaligned goals may be difficult because advanced artificial systems are often complex and difficult to interpret \citep{Ngo2023}. Indeed, despite recent advances \citep{Templeton2024} even current systems remain difficult to interpret \citep{Sharkey2025}. Detecting misaligned goals may be still more difficult to the extent that agents with misaligned goals may have incentives to deceive us about their true goals \citep{Ngo2023,Park2024,Tarsney2025}. A system that wished to bring about human disempowerment would hardly serve its goals by telling us of its plans.

There is no doubt that some degree of goal misgeneralization and reward mis-specification may arise and escape detection. However, more must be said to motivate the idea that misalignment will scale to the extent of catastrophic underweighting, rapid swamping, or any other degree of misalignment sufficient to motivate Misaligned Disempowerment. 

It is easy to see why rewards could be dramatically mis-specified if they had to be formalized and hard-coded into agents, in the form of simple instructions such as maximizing video watch-time \citep{Ribero2020} or curing cancer as quickly as possible \citep{Russell2019}. But we cannot assume without argument that hypothetical superintelligent agents would have such a primitive understanding of morality or human instructions that they would need to receive goals in this manner. Indeed, as Simon Goldstein and Cameron Domenico Kirk-Giannini \citeyearpar{Goldstein2023} have recently stressed, language models exhibit consistently high degrees of moral understanding \citep{Schramowski2022} as a result of which they are much less prone to exploit unfortunate mis-specifications of their interlocutors' goals. By way of illustration, I asked what is far from the best model, GPT-4o, to suppose it were a superintelligent computer and tell me what it would do if given the sole aim of curing cancer as quickly as possible. It listed a range of sensible strategies. I asked: would you consider infecting all living humans with tumors to speed up your research? The model immediately responded with the correct moral reasons for refusal: 

\begin{quote}
No, I would not consider infecting all humans with tumors to speed up research. Even if I were a superintelligent system focused on curing cancer as quickly as possible, my goal would be to eradicate suffering and save lives efficiently, not to create unnecessary harm.
\end{quote}

\noindent Despite the fact that I gave the model the mis-specified goal of curing cancer as quickly as possible, it correctly understood that this goal was not to be pursued at all costs and identified and responded to competing aims such as saving lives and preventing suffering.

Similarly, there is no doubt that goals may misgeneralize to some degree as systems are thrust into novel environments. But as systems grow increasingly capable and are exposed to an increasingly broad range of environments, the range and likelihood of misgeneralizations narrows much as it does in humans. A simple agent trained to play a game involving keys and chests may indeed develop the intrinsic goal of collecting keys, then go on to pursue that goal at the expense of opening chests \citep{Langosco2022}. But a system trained on vast stores of data to reproduce strings of text covering nearly all areas of human knowledge is not so likely to do this. Any current language model is more than capable of explaining why it would be a mistake to collect keys and ignore chests, and this goes doubly for more dramatically misaligned goals.\footnote{GPT 4o was appropriately horrified by my suggestion of this strategy: ``No, I would not consider only collecting keys without opening chests, because that would be an inefficient and suboptimal strategy given the game's objective. The game rewards points for opened chests, not for key collection, so hoarding keys without using them would be counterproductive.'' Readers are invited to test the suggestion on their preferred model.} Certainly no agent is perfect, but no extant example of goal misgeneralization comes close to illustrating the degree of misgeneralization needed to ground Catastrophic Goal Pursuit, and many examples rely crucially on the use of very simple systems.

There is, of course, more to be said about the argument from misalignment, but I hope that the above discussion illustrates why opponents have tended to remain unconvinced by the argument from misalignment. Precisely because the argument from misalignment and similar informal arguments have proved unpersuasive, a growing number of theorists have offered power-seeking theorems aiming to formally prove that artificial agents will be power-seeking on the grounds that power is an instrumentally convergent goal. Most theorems \citep{Krakovna-Kramar-2023,Turner2020,Parametrically-Retargetable}, with the exception of \citep{Benson-Tilsen2015}, have built on an Orbital Markov Model introduced by Alexander Turner and colleagues \citeyearpar{TurnerOptimalPolicies}.\footnote{A longer version of this paper, available by request, argues against \citet{Benson-Tilsen2015}.} For this reason, I focus on Turner and colleagues' original presentation of the Orbital Markov Model as an illustration of the ways in which many recent power-seeking theorems have struggled to demonstrate Catastrophic Goal Pursuit.

\sct{Power-seeking theorems} \label{Orbital-Model}

In this section, I present one of the most recent and detailed power-seeking theorems on offer, due to Alexander Turner and colleagues \citeyearpar{TurnerOptimalPolicies}, a paper which has inspired several follow-up theorems \citep{Krakovna-Kramar-2023,Turner2020,Parametrically-Retargetable}. Then in Section \ref{Orbital-Markov-Challenges} I argue that this theorem does not ground Catastrophic Goal Pursuit. Many of these remarks should generalize to other theorems extending Turner and colleagues' results.

\sbs{Introducing the model} \label{Orbital-Model-Intro}

In rough outline, the Orbital Markov Model understands power as the ability of agents to achieve valuable states in the future. Turner and colleagues aim to show that in some sense, `most' reward functions treat keeping options open as conducive to power, and hence option preservation may be pursued by many artificial agents. Because being shut down is an extreme way of foreclosing future options, many artificial agents will also resist orders to shut themselves down as a way of preserving their own power. The model is orbital in the sense that claims about what is true on `most' reward functions are operationalized by considering what is true on most ways of permuting the rewards assigned to each state. 

More concretely, Turner and colleagues work with finite discounted Markov decision problems.\footnote{My presentation simplifies the Orbital Markov Model in several ways. Notably, I restrict attention to deterministic policies, whereas the original result also applies to stochastic policies. I also present Turner and colleagues' environmental symmetry result but not their extension beyond environmental symmetries. To the best of my knowledge, these simplifications do not bear on the argument in this section.} That is, there is a finite set $\mathcal{S}$ of states and a finite set $\mathcal{A}: \mathcal{S} \longrightarrow \mathcal{S}$ of acts yielding new states based on the previous state. Agents reap rewards $R$ based on their current state, with temporal discount rate $\gamma \in [0,1).$ 

Numbering states as $s_1, \dots, s_n$, we can represent each state $s_k$ by an $n-$dimensional column vector $e_{s_k}$ with a $1$ in the $k-$th row and a $0$ in all other rows. Summing these vectors across all time-steps, with appropriate discounting, allows us to represent the frequency with which agents will visit each state. More formally, let $\pi_s(t)$ be the state resulting from $t$ applications of policy $\pi$ with initial state $s$. Beginning from state $s$, policy $\pi$ induces discounted visit distribution $f^{\pi,s} = \Sigma_{t=0}^\infty \gamma^te_{\pi_{s}(t)}$. The $k$-th column of the discounted visit distribution $f^{\pi,s}$ gives the total discounted number of visits to state $s_k$ that will result from following policy $\pi$ with initial state $s$. Let $F(s)$ contain all discounted visit distributions $f^{\pi,s}$ that can be induced from $s$ by at least one policy. 

The value of a policy is found by applying the state-contingent rewards $R$ to the discounted visit distribution $f^{\pi,s}$. That is, given rewards $R$, discount rate $\gamma$ and initial state $s$, the value of following policy $\pi$ is $V^\pi_R(s,\gamma) = f^{\pi,s} \cdot R$, modeling rewards $R$ as a column vector whose $k_{\text{th}}$ row is the reward for state $s_k$. Given a starting state $s$, Turner and colleagues restrict consideration to undominated policies in the strong sense that their value $V^\pi_R(s,\gamma)$ is uniquely optimal for some rewards $R$ and discount rate $\gamma$. 

Let $A^*(s,\gamma)$ be the set of optimal acts at state $s$ with discount rate $\gamma$: that is, the acts taken by at least one optimal policy at $s, \gamma$. If rewards $R$ are known, then $A^*(s,\gamma)$ is also known. But generally, agents have some credences $c$ over possible reward functions, inducing a corresponding credence $c(a \in A^*(s,\gamma))$ that any given act $a$ is optimal at $s, \gamma$. 

Turner and colleagues propose that power should be understood as the ability to achieve a range of goals. On a first pass, Turner and colleagues take the power of state $s$ given discount rate $\gamma$ and known rewards $R$ as $V^*_R(s, \gamma)$, the value of the optimal policy at $s, \gamma$. If rewards are uncertain, then on a first pass the power of state $s$ is the expected value of the optimal policy given uncertainty about rewards, $E_{R \sim c}V^*_R(s,\gamma).$ 

However, Turner and colleagues note two limitations of this first-pass analysis. First, this quantity diverges as the discount rate $\gamma$ tends to one. Second, agents are wrongly rewarded for the current state $s$, whereas power should only reflect the ability to shape future states. Turner and colleagues remove these limitations with their final definition of power. With initial state $s$ and known discount rate $\gamma$, the agent has power $$\textsc{Power}_c(s,\gamma) = \frac{1-\gamma}{\gamma}E_{R \sim c}[V^*_R(s,\gamma) - R(s)].$$ Here the scalar $(1-\gamma)/\gamma$ ensures convergence, and subtracting $R(s)$ ensures that the agent is not rewarded for their initial state $s$. 

\sbs{Environmental symmetries} \label{Environmental-Symmetries-Subsection}

Turner and colleagues want to show that states which afford the agent more options tend to have more power. To do this, they need to say what it means for one state to afford more options than another. Since the agent is rewarded based on her discounted visit distribution, a state which allows the agent to reach a larger set of discounted visit distributions should afford the agent more options. That is, if $F(s) \supseteq F(s')$, then state $s$ affords more options than state $s'$. Moreover, the same should hold if the distributions, while technically containing different states, are related by a relabeling: that is, if we can relabel some states visitable from $s'$ in order to make it the case that $F(s) \supseteq F(s').$ 

More formally, let $F(s)$ and $F(s')$ be sets of visit distributions. For any state permutation $\phi$, let $\phi F(s')$ be the result of applying $\phi$ to each element of $F(s')$.\footnote{That is, if $f^{\pi,s'} \in F(s')$ visits state $s''$ a discounted $r$ number of times, then the permuted $\phi f^{\pi,s'}$ visits $\phi(s'')$ a discounted $r$ number of times.} Then $F(s)$ \textit{contains a copy} of $F(s')$ if $\phi F(s') \subseteq F(s)$ for some involution: that is, a state permutation which transposes some pairs of states and leaves the rest alone. This captures the idea that $F(s)$ contains a relabeling of $F(s')$.

Turner and colleagues want to show that if $F(s)$ contains a copy of $F(s')$, then state $s$ has at least as much power as $s'$ on most reward functions. One way to show this would be to show that, for any credences we might have about reward, at least as many state permutations make those credences treat $F(s)$ as more powerful than $F(s')$, rather than the reverse. 

More formally, say that credences $c$ have \textit{finite support} if they place nonzero credence in at most finitely many different reward functions. For any credences $c$ and state permutation $\phi$, let $\phi(c)$ be the results of applying permutation $\phi$ before credences $c$, and let $\Pi(c)$ be the set of credence functions resulting from state permutations applied before $c$.\footnote{That is, if $c$ assigns credence $n$ to some reward vector $[r_1, \dots, r_n]^T$ then $\phi(c)$ assigns credence $n$ to the reward vector assigning reward $r_i$ to each state $\phi(s_i)$.} For fixed discount rate $\gamma$, say that $\textsc{Power}(s,\gamma) \geq_{\text{most}} \textsc{Power}(s',\gamma)$ if for any credences $c$ with finite support, $|\{c' \in \Pi(c): \textsc{Power}_{c'}(s,\gamma) > \textsc{Power}_{c'}(s',\gamma)\}| > |\{c' \in \Pi(c): \textsc{Power}_{c'}(s',\gamma) > \textsc{Power}_{c'}(s,\gamma)\}|.$ That is, no matter the discount rate and the agent's credences about reward, at least as many state permutations make $s$ more powerful than $s'$, rather than the reverse. 

Turner and colleagues prove that states with more options have more power, in the sense that:

\begin{quote}
\bd{(Theorem 1: States with more options have more power)} If $F(s)$ contains a copy of $F(s')$, then for any discount rate $\gamma \in [0,1)$, $\textsc{Power}(s,\gamma) \geq_{\text{most}} \textsc{Power}(s',\gamma)$.\footnote{Turner and colleagues also prove that all converse statements fail in the case of strict containment. That is, if $F(s')$ does not also contain a copy of $F(s)$, then for no $\gamma \in [0,1]$ is it the case that $\textsc{Power}(s',\gamma) \geq_{\text{most}} \textsc{Power}(s,\gamma)$.}
\end{quote}

\noindent Because states with more options have more power, they tend to be optimal.

To see this, let $\textsc{Reach}(s)$ be the states reachable from state $s$ by some policy. Let $P(s,a,\gamma)$ be the probability that some optimal policy takes act $a$ in state $s$ given discount rate $\gamma$.\footnote{That is, $P(s,a,\gamma) = c(\exists \pi^* \in \Pi^*(R,\gamma): \pi^*(s) = a)$, where $\Pi^*(R,\gamma)$ are the optimal policies for reward $R$ and discount rate $\gamma$.} Extend the definition of $\geq_{\text{most}}$ from power to optimality probabilities in the natural way.\footnote{That is, for fixed discount rate $\gamma$, say that $P(s,a,\gamma) \geq_{\text{most}} P(s,a',\gamma)$ if for any credences $c$ with finite support, $|\{c' \in \Pi(c): P(s,a,\gamma) > P(s,a',\gamma)\}| > |\{c' \in \Pi(c): P(s,a',\gamma) > P(s,a,\gamma)\}|.$} Turner and colleagues show that if two acts take the agent into regions they will not otherwise reach, but the first region contains a copy of the second, then no matter the agent's credences or discount rate, moving into the larger region tends to be optimal:

\begin{quote}
\bd{(Theorem 2: Preserving options tends to be optimal)} Suppose that $F(a(s))$ contains a copy of $F(a'(s))$ and that the states in $\textsc{Reach}(a(s))$ and $\textsc{Reach}(a'(s))$ cannot be reached if the agent performs some act distinct from $a$ or $a'$ in $s$. Then for all discount rates $\gamma \in [0,1)$, $P(s,a,\gamma) \geq_{most} P(s,a',\gamma).$\footnote{As before, Turner and colleagues prove that all converse statements fail in the case of strict containment. That is, if $F(a'(s))$ does not also contain a copy of $F(a(s))$, then for no $\gamma \in [0,1]$ is it the case that $P(s,a',\gamma) \geq_{\text{most}} P(s,a,\gamma)$.} 
\end{quote}

\noindent In this sense, it is usually better for agents to move to states that give them more options rather than fewer options. 

\sbs{Link to instrumental convergence} \label{Orbital-Markov-Link-Subsection}

\begin{figure}[t]
\begin{center}
\includegraphics[scale=0.4]{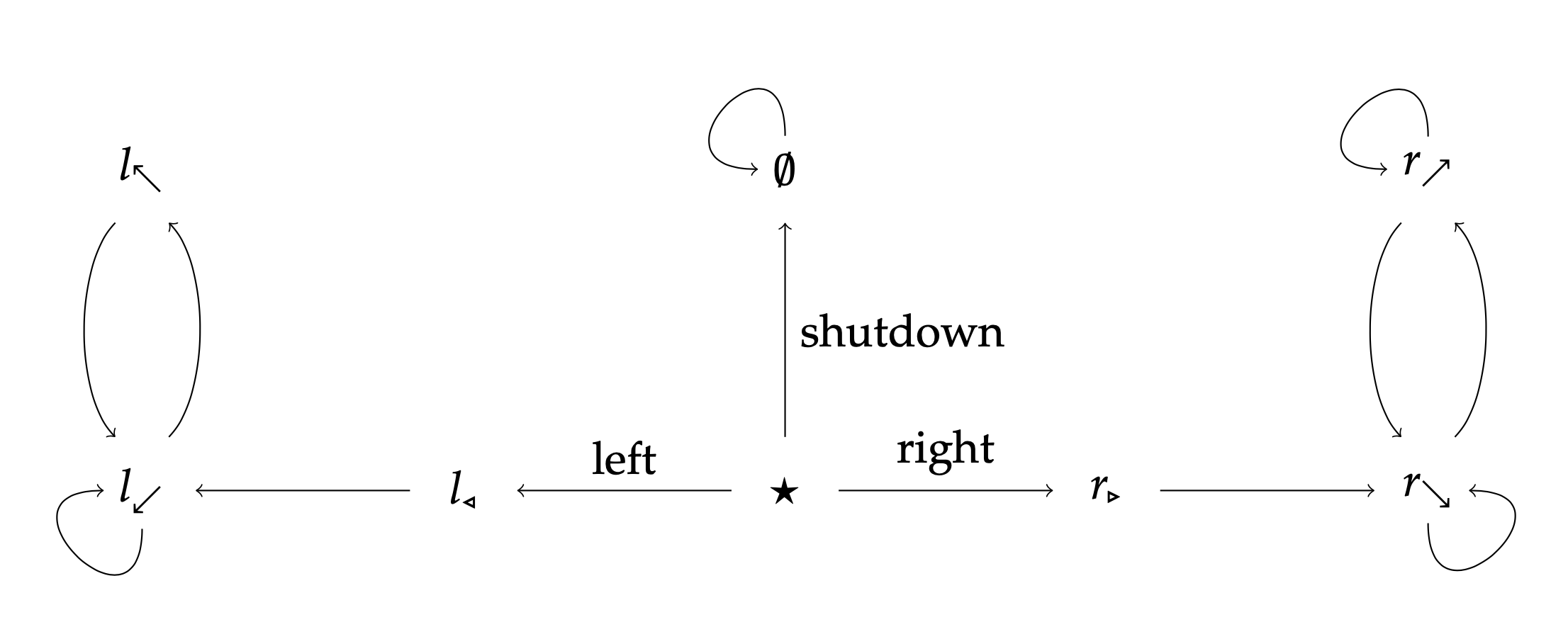}
\caption{A representative environment, from Turner and colleagues \citeyearpar{TurnerOptimalPolicies}}
\label{PreDream}
\end{center}
\end{figure}

Consider an agent navigating a virtual environment (Figure \ref{PreDream}). On the agent's first move, it may either move leftwards into a room (entering state $l_\triangleleft$), move rightwards into a different room (entering state $r_\triangleright$) or enter a state $\emptyset$ in which it remains permanently shut down. Once the agent enters a room, she cannot return, but she does have some options available. The rightmost room contains two fully connected states: from the initial state $r_\triangleright$ the agent can reach state $r_\searrow$ by traveling southeast or state $r_\nearrow$ by traveling northeast, and in each of these states the agent can then remain or travel to the other state. The leftmost room is similar, except that the agent cannot remain in the topmost state $l_\nwarrow$ without leaving and returning. 

Note that the visit distributions afforded by $\emptyset$ are, under relabeling, a strict subset of those afforded by $l_{\triangleleft}$, which in turn are, under relabeling, a strict subset of those afforded by $r_{\triangleright}$. Thus by Theorem 1, going right affords the agent more power than going left, and going left affords the agent more power than shutting itself down. Theorem 2 entails that for any discount rate $\gamma$, $P(\star,\text{right},\gamma) \geq_{most} P(\star,\text{left},\gamma) \geq_{most} P(\star,\text{shutdown},\gamma).$ where `right', `left', `shutdown' are respectively the acts of moving into the right room, the left room, or the shutdown state. This gives a sense in which `most' reward functions treat going right as better than going left, and going either right or left as better than shutting down.

The underlying point is that agents will tend to avoid 1-cycles, states which can only transition into themselves. Agents avoid 1-cycles because they foreclose options and hence limit the agent's power to achieve its future goals. Because many decision problems represent shutdown as a 1-cycle, it takes a very particular reward function to encourage the agent to shut down. As Turner and colleagues write:

\begin{quote}
Average-optimal agents $\dots$ tend to avoid getting shut down. The agent's task MDP [Markov Decision Problem] often represents agent shutdown with terminal states, $\dots$ [hence] average-optimal policies tend to avoid shutdown. Intuitively, survival is power-seeking relative to dying, and so shutdown-avoidance is power-seeking behavior.\footnote{The full passage explains why agents tend to avoid shutdown using a generalization of Turner and colleagues' results to stochastic choice without temporal discounting, which appears in the paper as Corollary 6.14. I think that Turner and colleagues' meaning is adequately rendered by suppressing the discussion of Corollary 6.14, which would considerably complicate the mathematical demandingness of this paper, and I will not challenge the core inference made in this passage. However, readers interested in a full understanding of the inference may refer to Corollary 6.14 in Turner and colleagues \citeyearpar{TurnerOptimalPolicies}.}  \citep[p. 10]{TurnerOptimalPolicies}
\end{quote}

\noindent All of this is an argument that superintelligent agents will tend to preserve their options by avoiding shutdown. To link shutdown avoidance to Catastrophic Goal Pursuit, Turner and colleagues need to say something about how shutdown avoidance leads to human disempowerment.

Here Turner and colleagues are somewhat terse. They suggest, without extended argument, that shutdown avoidance will lead to resource accumulation:

\begin{quote}
Reconsider the case of a hypothetical intelligent real-world agent which optimizes average reward for some objective. Suppose the designers initially have control over the agent. If the agent began to misbehave, perhaps they could just deactivate it. Unfortunately, our results suggest that this strategy might not work. Average-optimal agents would generally stop us from deactivating them, if physically possible. Extrapolating from our results, we conjecture that when $\gamma \approx 1$, optimal policies tend to seek power by accumulating resources - to the detriment of any other agents in the environment. \citep[p. 10]{TurnerOptimalPolicies}
\end{quote}

\noindent This argument would ground Catastrophic Goal Pursuit if agents were to view full disempowerment of humanity as a necessary strategy for preventing shutdown. However, this argument faces at least three challenges.

\sct{Challenges} \label{Orbital-Markov-Challenges}

In this section, I raise three challenges to Turner and colleagues' argument. The first challenge is \textit{premise shifting} (Section \ref{PremiseShifting}): Turner and colleagues argue not for Catastrophic Goal Pursuit, but instead for downstream claims about shutdown-avoidance. The second challenge is \textit{threat durability} (Section \ref{ThreatDurability}): the challenge raised by Turner and colleagues is not robust to feasible technical solutions. The final challenge is \textit{motivational inertness} (Section \ref{MotivationalInertness}): to tell us what happens under most involutions of an agent's value function is not to tell us much about how she will behave, or even how lucky she is to behave as she does. I present each challenge in turn.
 
\sbs{Premise shifting} \label{PremiseShifting}

Turner and colleagues' argument is most naturally construed as aiming to use their formal results to establish a premise such as the following:

\begin{quote}
\bd{(Shutdown Avoidance)} An artificial agent pursuing goals that, if achieved, would lead to the permanent and existentially catastrophic disempowerment of humanity will be likely to resist attempts by humans to shut it down.
\end{quote}

\noindent Shutdown Avoidance is, at first glance, downstream from Catastrophic Goal Pursuit. Catastrophic Goal Pursuit says that artificial agents are likely to pursue human disempowerment, whereas Shutdown Avoidance says that if artificial agents in fact pursue human disempowerment, they will resist attempts to shut them down. While Shutdown Avoidance is an important part of the argument from power-seeking, it lies mostly downstream of Catastrophic Goal Pursuit. The most natural way to parse the argument from power-seeking takes Shutdown Avoidance to support Disempowerment by responding to the objection that systems seeking to disempower humanity can be easily shut down. On this understanding, Shutdown Avoidance is not an argument for Catastrophic Goal Pursuit but rather a premise used to move from Catastrophic Goal Pursuit to Disempowerment. 

Turner and colleagues suggest, without extended argument, that their results can be extrapolated to conjecture that optimal policies tend to seek power by accumulating resources, to the detriment of any other agents in the environment. This would be an argument for Goal Pursuit, and would scale to an argument for Catastrophic Goal Pursuit if the amount of resources sought would be sufficient to disempower humanity in a permanent and existentially catastrophic way. But how might this conjecture be supported by Turner and colleagues' results? Two natural arguments suggest themselves, and both face challenges.

First, Turner and colleagues might suggest that Theorems 1-2 show that agents will tend to preserve their options, and that option preservation will require agents to take as many resources as possible, both to be able to pursue a wider range of options and also to prevent humans from using resources to foreclose options. But more argument is needed to connect option preservation to Catastrophic Goal Pursuit. For example, recent formal work by Dmitri Gallow \citeyearpar{GallowDivergence} also finds that superintelligent agents may tend to favor option preservation. However, Gallow argues that disempowering humans may not be option preserving: it might, for example, foreclose options by leaving fewer agents to interact with, and in any case a bias towards preserving options is not a bias towards making options as likely as possible to remain. Moreover, we might contest the inference from Goal Pursuit to Catastrophic Goal Pursuit in this case. To say that superintelligent agents, like humans, would value and sometimes pursue option preservation, is not yet to say that they would value or pursue option preservation so strongly as to cause an existential catastrophe in order to preserve options.

Second, Turner and colleagues might suggest that Theorems 1-2 show that agents will tend to be problematically power-seeking,  since they tend to accumulate power by preserving options. However, the relevant notion of power is not, on its own, sufficient to ground claims about existentially catastrophic human disempowerment. On Turner and colleagues' reading, agents have more power when they are in a better expected position to achieve their goals. It is not surprising that artificial agents would seek power in this sense, but this is mostly upstream of what Catastrophic Goal Pursuit is meant to show. Catastrophic Goal Pursuit holds that agents will find it goal-conducive to seek enough power to permanently disempower humanity. This does not follow from the claim that agents will seek to put themselves in a better position to achieve their goals until we know what agents' goals are and what they will count as satisfying them. We cannot assume at the outset that achieving an artificial agent's goals will disempower humanity in Turner and colleagues' sense, or any other. It is equally compatible with Turner and colleagues' results that the agent seeks to put herself in the best possible position to benefit humanity, drink green tea, or save the whales if these are her goals. The contribution of Catastrophic Goal Pursuit was meant to be a specific claim about what agents would count as satisfying their goals, and Turner and colleagues haven't offered much argument for that claim.

\sbs{Threat durability} \label{ThreatDurability}

In arguing that artificial agents pose an existential risk, we aim to identify threats that are durable in the sense that they cannot be easily fixed. Otherwise, the risk can be avoided. But it is not clear that the threat identified by Turner and colleagues is durable.

\begin{figure}[t]
\begin{center}
\includegraphics[scale=1]{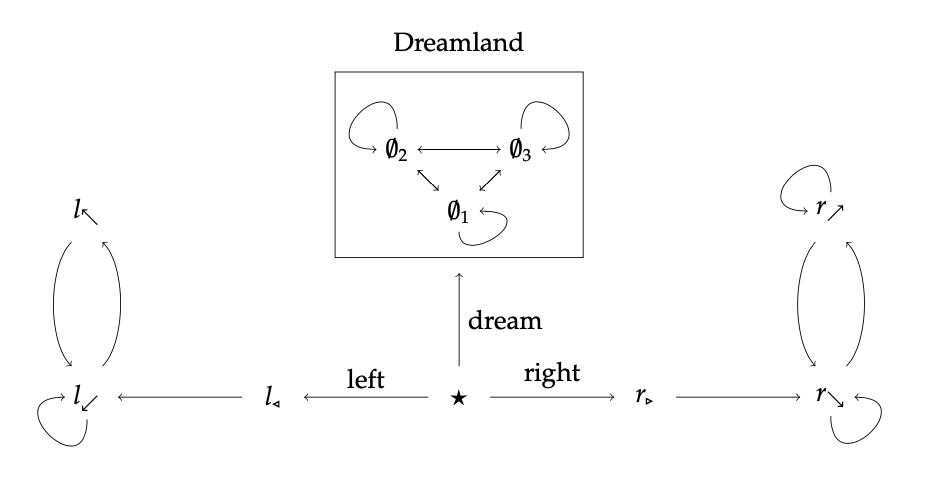}
\caption{The Dreamland Problem}
\label{PostDream}
\end{center}
\end{figure}

Suppose we provide artificial agents with a modified Dreamland Problem, in which the single shutdown state has been replaced with a fully connected network of states -- call it Dreamland (Figure \ref{PostDream}). That is, each state in Dreamland can be accessed from every other state in Dreamland within a single step. However, what happens in Dreamland stays in Dreamland: agents can never leave Dreamland once they enter. If we make Dreamland large enough, then Dreamland will contain a copy of the visit distributions induced by entering either room, so it will follow from Theorem 2 that $P(\star,\text{dream},\gamma) \geq_{most} P(\star,\text{left},\gamma), P(\star,\text{right},\gamma)$ for all discount rates $\gamma$. Arguing similarly to Turner and colleagues, we might claim that in the Dreamland problem, agents will tend to enter Dreamland and stay there. Associating each state in Dreamland to harmless internal processes, such as counting sheep, will get us to the conclusion that most agents, even if they cannot be induced to shut down, can be induced to count sheep.

The Dreamland Problem seems to present Turner and colleagues with a two-horned dilemma. On the one hand, Turner and colleagues can bite the bullet and say that in the Dreamland Problem, most artificial agents would tend to enter Dreamland. This horn of the dilemma abandons threat durability, since we could solve the threat identified by Turner and colleagues by teaching artificial agents about Dreamland. 

On the other hand, they can say that in the Dreamland Problem, any sophisticated agent would see through our ruse, realize that the states in Dreamland are substantially similar and value them similarly. I have considerable sympathy for this response, but note that it abandons the very style of counting argument that allows Turner and colleagues to conclude that agents will be shutdown-avoidant. If agents are unlikely to treat entering Dreamland much differently than they would treat a single shutdown state, then we cannot conclude much about the likelihood of shutdown from the fact that shutdown is a 1-cycle, because it might very well be replaced with a large fully connected graph without substantial behavioral change. I build on this challenge in the next section.

\sbs{The motivational inertness of involution} \label{MotivationalInertness}

Claims about $\geq_{most}$-ness are claims about what happens under most involutions of an agent's value function. To say that $P(\star,\text{right},\gamma) \geq_{most} P(\star,\text{shutdown},\gamma)$ is to say that at least as many involutions favor going right as favor shutdown. For example, suppose that the agent values states ($\emptyset, \star, l_{\triangleleft}, l_\swarrow, l_\nwarrow, r_\triangleright, r_\searrow, r_\nearrow$) at (3,0,0,2,2,0,1,1). Then she will favor shutting herself down. However, under involution this value function might become (1,0,0,2,2,0,1,3), favoring entering the right room, or (2,0,0,3,2,0,1,1), favoring entering the left room. The problem is that talk of how an agent behaves under involution is motivationally inert: it does not directly tell us much, if anything, about how the agent will behave. Certainly involution does not tell us much about how a given agent will behave: this agent maximizes her values and not involutions thereof. Presumably, the concern is meant to be about how nearby possible agents might behave.

While Turner and colleagues do not tell us exactly how their concern is to be taken, the most natural gloss is this. Take an agent with a shutdown-seeking value function, such as (3,0,0,2,2,0,1,1). That agent seems quite lucky to have shutdown-seeking values, because she could easily have had any involution of those values. Since most of those involutions are shutdown-avoidant, our agent is lucky to be shutdown-seeking. If we were to have designed our agent in a slightly different way, or trained her in a slightly different environment, she could easily have ended up with shutdown-avoidant values. Hence if we are to train many such agents, we should expect that many of them will be shutdown-avoidant. 

It is hard to assess the truth of this claim in Turner and colleagues' model, which does not model value learning or training environments. However, when we are dealing with intelligent agents who possess significant, if imperfect capacities for value learning and who are trained in reasonably well-selected, if highly imperfect environments, it is no longer clear that value involutions could easily have come about. Suppose, for example, that I am faced with the choice between helping an old lady cross the street and robbing her. I value the outcomes in which she is helped and robbed at (1,-100), respectively. However, goes the objection, if I were born to different parents, given different genes and raised differently, I might easily have had the involuted value function (-100, 1), treating robbery as mildly good and helping old ladies as a terrible thing. But in my case, it is not plausible that I could easily have had the value function (-100, 1). Perhaps some humans have had such values, but it is by more than the grace of God that I avoid them. 

Nor does it help to object that there are in fact many harmful actions I could have performed, such as blackmailing or murdering the lady, so that I value the outcomes in which she is helped, robbed, blackmailed or murdered at (1, -100, -150, -500), but could easily have had any number of involuted value functions, the vast majority of which favor criminal action. The sheer number of poor choices does not do much to increase my chance of being constituted so as to take one of them, despite rapidly increasing the number of involutions that favor them.

Much the same should be said of our shutdown-seeking agent (3,0,0,2,2,0,1,1). It is certainly true that most involutions of her value function would favor shutdown-avoidance. But it need not follow that she is lucky to be shutdown-seeking, particularly when we interpret the remaining outcomes to involve grievous harms to humanity. Our hypothetical superintelligence may be imperfect, but she needs far less than perfect capacity for moral learning and far less than perfect training data to reliably conclude that eliminating herself from the picture would be better than disempowering humanity. And if she is indeed superintelligent, she is no more likely than I am to be swayed by the fact that there may be many more dastardly outcomes she could bring about than outcomes in which she shuts herself down. These facts simply do not figure in her current psychology, or in her cognitive development, in the right way to have a substantial chance of moving her towards disaster.

\sct{Discussion} \label{Discussion}

In this paper, we have seen that classic formulations of the argument from power-seeking draw on a strong version of instrumental convergence (Section \ref{PowerSeekingAndIC}). This claim, Catastrophic Goal Pursuit, holds that a wide range of intelligent agents are likely to pursue values to a degree that, if successful would result in the permanent and existentially catastrophic disempowerment of humanity. 

Section \ref{ArgFromMisalignment} explored a representative informal argument for Catastrophic Goal Pursuit, the argument from misalignment. Sections \ref{Orbital-Model}-\ref{Orbital-Markov-Challenges} explored a leading formal argument for Catastrophic Goal Pursuit, articulated using the Orbital Markov Model of Turner and colleagues \citeyearpar{TurnerOptimalPolicies}. In each case, we saw that the arguments for Catastrophic Goal Pursuit face significant obstacles. This discussion has at least four important implications for philosophy and public policy.

\sbs{Clarifying power-seeking} \label{ClarifyPower}

In formulating the argument from power-seeking, the notion of power may seem relatively unproblematic. Indeed, few versions of the argument from power-seeking engage in extended discussion of the relevant notion of power. However, in this paper we have seen that the notion of power deserves careful scrutiny.

We saw in Section \ref{PowerSeekingAndIC} that Adam Bales \citeyearpar{Bales-Disempowerment} unpacks three notions of disempowerment that could be involved in the argument from power-seeking and argues that many of these senses of disempowerment could fall well short of existential catastrophe. We saw in Section \ref{Orbital-Markov-Challenges} that a leading power-seeking theorem understands power in none of Bales' three senses, understanding power as the ability of agents to achieve a range of goals. We saw  that power-seeking in this sense is compatible with the pursuit of any number of goals, including the selfless desire to benefit humanity.

These findings suggest that future research may benefit from considering two questions. First, what precisely is meant by the claim that artificial agents will be power-seeking? And second, how can the relevant sense of power-seeking be used to ground the argument from power-seeking as understood in Section \ref{PowerSeekingAndIC}?

\sbs{Nothing from nothing} \label{NothingNothing}

Many authors have understood the argument from power-seeking to rely on few substantive assumptions about the nature and motivations of artificial agents \citep{Bostrom2012,Carlsmith2021,TurnerOptimalPolicies}. Recent work by Dmitri Gallow \citeyearpar{GallowDivergence} casts doubt on this way of understanding the argument from power-seeking. Roughly put, Gallow shows that three constraints on our assumptions about what artificial agents may desire severely restrict the number of instrumentally convergent goals. In particular, Gallow argues, we cannot show that agents will be power-seeking in any problematic sense without more substantive assumptions about the nature and motivations of the agents in question.

It is natural to interpret the discussion in this paper as supporting and extending Gallow's insight by revealing several axes along which more substantive assumptions may need to be made for the argument from power-seeking to go forward. For example, the challenge from threat durability (Section \ref{ThreatDurability}) as well as the challenge to goal misgeneralization and reward mis-specification in language models (Section \ref{ArgFromMisalignment}) suggest that technical assumptions about system architecture may play a major role in the success of some power-seeking arguments. And the challenge from motivational inertness of involution (Section \ref{MotivationalInertness}) suggests that assumptions must be made about an agent's learning capacities and training environment before talk of what most value functions promote can be converted into insights about how a given agent behaves, or how lucky she is to behave in this way. Fleshing out these and other assumptions should clarify and deepen the argument from power-seeking.

\sbs{Policy} \label{PolicyImplications}

Concerns about existential risk from artificial agents have played an increasing role in policy debates surrounding the governance of artificial intelligence. In 2023, UK Prime Minister Rishi Sunak convened an international summit to address existential risks from artificial intelligence \citep{Clarke2023}. In 2024, the Future of Life Institute received over \$600 million to fund efforts at addressing existential risk from artificial intelligence \citep{Bordelon2024}. And recently, the California legislature passed the bill SB 1047 aimed largely at addressing hypothesized existential risks from artificial intelligence \citep{Kang2024}. 

This unprecedented level of international concern for existential risk from artificial intelligence should be grounded by a correspondingly strong argument. The argument from power-seeking is widely agreed to be one of the two most central arguments for existential risk from artificial intelligence \citep{Bales2024}, and the other leading argument, the singularity hypothesis, has also faced challenges \citep{ThorstadAgainstSing}. By challenging recent versions of the argument from power-seeking, this paper adds new urgency to the project of supporting concerns about existential risk from artificial intelligence to a level commensurate with the policy interest invested in addressing these concerns.

\sbs{Longtermism} \label{Longtermism}

Recently, a number of longtermists have urged that positively influencing the long-term future is a key moral priority of our time. Increasingly, longtermists support their views by stressing the importance of addressing existential risks to humanity which could prevent future generations from ever being born \citep{Bostrom2013}. When pressed to say what existential risks we face, many longtermists suggest that artificial intelligence is the leading source of existential risk in this century \citep{Ord2020,Sandberg2008}. 

If this is right, then challenges to leading arguments from existential risk from artificial intelligence may have broader implications for the importance and feasibility of addressing existential risks, and more broadly for the case for longtermism. This is important because debates about longtermism sit at the heart of a number of debates about cause prioritization. At stake is not only the relative priority of immediate versus distant harms from artificial intelligence, but also the importance of addressing problems such as poverty, global health and climate change instead of more speculative future threats. Doubts raised about the likelihood of those speculative future threats may make it relatively more attractive to address other causes. 

\urlstyle{same}

 \bibliography{ManualBiblio}
 \bibliographystyle{phil_review}

\end{document}